\documentclass[letterpaper, 10 pt, conference]{ieeeconf}  

\IEEEoverridecommandlockouts                              

\usepackage{epsfig} 
\usepackage{mathptmx} 
\usepackage{times} 
\usepackage{amsmath} 
\usepackage{amssymb}  
\usepackage{color}
\usepackage{xcolor}
\usepackage{preambles}
\usepackage{amsmath}
\usepackage{graphicx}
\usepackage{url} 
\usepackage[format=plain,labelsep=period]{caption}
\usepackage{wrapfig}
\usepackage[list=on,listformat=simple]{subcaption}
\usepackage{hyperref}
\usepackage{url}
\usepackage{caption}
\usepackage{subcaption}
\hypersetup{colorlinks = true, citecolor=black}

\usepackage{tikz}
\let\oldmarginpar\marginpar
\renewcommand{\marginpar}[2][rectangle,draw, text=white,text width= 2cm,rounded corners]{
    \oldmarginpar{
    \scriptsize \tikz \node at (0,0) [#1]{#2};}
    }

\title{\LARGE \bf
Resolved Motion Control for 3D Underactuated Bipedal Walking using Linear Inverted Pendulum Dynamics and Neural Adaptation}
\author{Victor C. Paredes$^{1}$ and Ayonga Hereid$^{1}$
\thanks{*This work was supported in part by the National Science Foundation under grant FRR-21441568. }%
\thanks{$^{1}$Mechanical and Aerospace Engineering, Ohio State University, Columbus, OH, USA. {\tt\footnotesize (paredescauna.1, hereid.1)@osu.edu.}}%
}

\usepackage[capitalise]{cleveref}

\usepackage[noadjust]{cite}

\DeclareMathOperator{\sech}{sech}

\begin{document}

\maketitle
\thispagestyle{empty}
\pagestyle{empty}

\begin{abstract}
We present a framework to generate periodic trajectory references for a 3D under-actuated bipedal robot, using a linear inverted pendulum (LIP) based controller with adaptive neural regulation. We use the LIP template model to estimate the robot's center of mass (CoM) position and velocity at the end of the current step, and formulate a discrete controller that determines the next footstep location to achieve a desired walking profile.
This controller is equipped on the frontal plane with a Neural-Network-based adaptive term that reduces the model mismatch between the template and physical robot that particularly affects the lateral motion. Then, the foot placement location computed for the LIP model is used to generate task space trajectories (CoM and swing foot trajectories) for the actual robot to realize stable walking. We use a fast, real-time QP-based inverse kinematics algorithm that produces joint references from the task space trajectories, which makes the formulation independent of the knowledge of the robot dynamics. Finally, we implemented and evaluated the proposed approach in simulation and hardware experiments with a Digit robot obtaining stable periodic locomotion for both cases.
\end{abstract}

\section{introduction}
Bipedal robotic locomotion is inherently an unstable process that requires proper planning of the body and swing foot trajectories to stabilize it. For instance, well-known methods such as the Divergent Component of Motion (DCM) \cite{englsberger2015three, griffin2016model}, the Zero Moment Point (ZMP) \cite{kajita20013d} or the Capture Point (CP) \cite{pratt2012capturability}, rely on the appropriate planning of the trajectory of the Center of Mass (CoM) and swing foot.
Other dynamic methods are based on the generation of whole periodic orbits that exhibit attractiveness \cite{goswami1996limit}. Many approaches are available to produce those trajectories, ranging from trajectory optimization \cite{hereid20163d, da2019combining}, template models based design using LIP \cite{gong2021one, xiong20213d}, Spring Loaded Inverted Pendulum \cite{xiong2020dynamic}, Centroidal Models \cite{xie2021glide} or reinforcement learning frameworks that learns stable trajectories \cite{castillo2020hybrid}. These methods are especially well suited when the system presents under-actuation and the trajectory planning must consider the passive or uncontrolled dynamics of the robot.  
Usually, the generation of a periodic orbit results in efficient dynamic walking gaits. However, in practice, these gaits require an additional stabilizing controller that adds robustness against model or terrain uncertainty and external disturbances \cite{paredes2020dynamic, castillo2020hybrid}. They usually use heuristic regulators that provide an intuitive human-inspired stabilizing strategy based on representative states of the robot, such as torso velocity and orientation~\cite{raibert1986legged}.

\begin{figure}
\centering
\vspace{2mm}
\includegraphics[width=0.7\columnwidth]{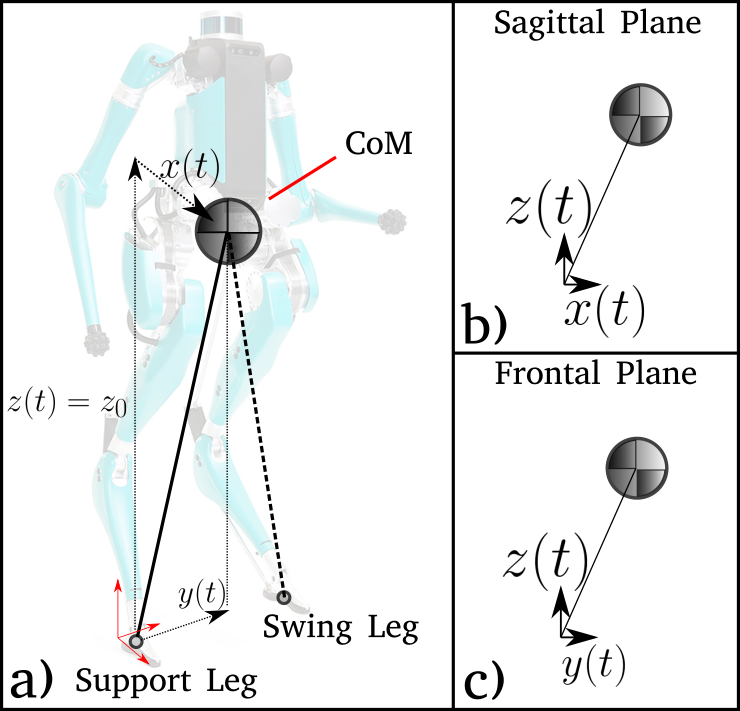}
\caption{\small{a) The linear inverted pendulum (LIP) is used to represent Digit with its total mass concentrated on the CoM and with massless legs. We divide the motion of the robot in: b) Sagittal Plane, to describe the forward and backward walking and the c) Frontal Plane, to describe the lateral motion.}} 
\label{fig:lip_digit}
\end{figure}



On the other hand, although the simplified template models have an imperfect representation of the actual robotic model, they capture the overall behavior of the robot with certain degree of accuracy and with a simple and analyzable model which has the advantage of providing a principled design approach for an stabilizing controller. 
Template based approaches developed in \cite{gong2021one, xiong20213d, teng2021safety} has produced stable periodic walking for the robot Cassie, a bipedal robot without torso and lightweight legs. Their approach uses a linear inverted pendulum as a template model which is used to produce a discrete stepping controller that converges to the LIP desired orbits. 
Since the LIP template model provides only an approximation, different robots might have different levels of mismatch with the predictions provided by the LIP. 

In this paper, we propose using a LIP template model to generate a stepping controller  equipped with an adaptive learning regulator to improve the stepping location that accounts for the mismatch between the template model and the actual robot. Note that the neural network learns the residual on the LIP model, and does not consider the model of the robot for the learning. On the other hand, model based Adaptive controllers such as L1 control methods \cite{nguyen20151} use the nominal knowledge of the robot dynamics. Similarly using learning to leverage the residual dynamics in real time as shown in \cite{sun2021online} use the robot's dynamics. However, in real-time applications, these model based controllers might require longer computation time than an inverse kinematics based trajectory generator, which in this work we highlight as a fast algorithm that can generate stable walking with a joint level PD control. 

Given the desired foot location we generate hand-crafted trajectories on task space which enforces the LIP conditions (constant height) and provides a suitable swing foot trajectory (enough foot clearance and soft impacts).  The task space trajectories are converted into desired joint level trajectories by running a QP-based inverse kinematics (QP-IK) algorithm in real-time. The joint references are tracked using a standard PD controller with feed-forward terms for torso orientation and robot height.
Consequently, our controller does not depend on the knowledge of the robot's dynamics but only its kinematics structure.   

In this work we use the robot Digit, a bipedal robot built by Agility Robotics, corresponding to the next iteration of Cassie. It that has a torso, arms and lightweight legs (See \figref{fig:lip_digit}).
The main contributions of this paper are:
\begin{itemize}
    \item Formulating an neural-adaptive framework that is based on the LIP model for foot placement based control. The adaptive term improve the gap between the LIP model and the actual Digit model.
\item The formulation of a task-space inverse kinematics solver that provides joint references in real-time.
\item The simulation and hardware experiments of the LIP-based controller with neural adaptation on the robot Digit.
\end{itemize}

The paper structure starts with \secref{sec:lip} that provides the mathematical background of the LIP model, including its phase portrait and a description of the forward and lateral walking. \secref{sec:traj} explains how to obtain a periodic trajectory based on the phase portrait of LIP and the symmetry of the gait. \secref{sec:control} formulates a discrete-time neural adaptive feedback control that stabilizes the step-to-step dynamics. \secref{sec:ik} presents the formulation of the task space objectives to achieve the foot placement provided by the stepping controller and a QP-based inverse kinematics algorithm for generating joint reference trajectories from the task-space trajectories. Finally, \secref{sec:results} presents the simulation and hardware experimental results of the proposed approach on the 3D biped robot, Digit.
\section{Linear Inverted Pendulum Model}
\label{sec:lip}

The LIP model is one of the most commonly used template models that captures the motion of the CoM dynamics of bipedal robots \cite{kajita20013d, pratt2012capturability, gong2021zero}. We use a 2D LIP model that provides a planar representation of motion that is linear and has only two states, the center of mass and its velocity with respect to the support foot, making it simple enough for stability analysis. Note that the support leg and the swing leg swap functions after impacting the ground, and in this work we consider that the swapping is instantaneous, providing only a single support phase (SSP). Our particular assumptions of the LIP model are:

\begin{itemize}
    \item The legs are massless, or in practice, they are lightweight.
    \item We consider that the robot has a point mass centered at the center of mass (CoM).
    \item The height of the CoM is constant.
    \item The robot has point-feet, or non-actuated ankles.
\end{itemize}

\subsection{LIP Dynamics}

Without loss of generality, we can choose the sagittal Plane (\figref{fig:lip_digit}) to analyze the LIP dynamics, which are represented by a linear second order system~\cite{kajita20013d}:
\begin{align}
\label{eq:lip-dynamics}
    \ddot{p}_x(t) = \frac{g}{z_0} p_x(t), 
\end{align}
where $p_x$ is the position of the CoM with respect to the support foot, $g$ is the gravity acceleration constant, and $z_0$ the constant CoM height. Let $\mathbf{x}(t) := (p_x(t),v_x(t))$ with $v_x(t) := \dot{p}_x(t)$ be the LIP states, then \eqref{eq:lip-dynamics} can be written in the state-space form:
\begin{align}
    \dot{\mathbf{x}} = \begin{bmatrix} 0 & 1 \\ \lambda^2 & 0 \end{bmatrix}  \mathbf{x} = A_{SSP} \mathbf{x},
\end{align}
where $\lambda = \sqrt{\frac{g}{z_0}}$.
Since the dynamics are linear, the states can be solved for a desired time $t$, given any set of initial conditions $\mathbf{x}_0=(p_{x}(0), v_{x}(0))$ as:
\begin{align}
    \mathbf{x}(t) = e^{A_{SSP} t} \mathbf{x_0} =    \underbrace{\begin{bmatrix} \cosh(\lambda t) & \frac{1}{\lambda} \sinh(\lambda t) \\
    \lambda \sinh(\lambda t) & \cosh(\lambda t) \end{bmatrix}}_{M(t)}     \mathbf{x_0},
     \label{eq:nonautonomous}
\end{align}



\begin{figure}
\centering
\vspace{2mm}
     \begin{subfigure}[b]{1\columnwidth}
         \centering
        \includegraphics[trim={0cm 0cm 0cm 0cm},clip,width=0.8\columnwidth]{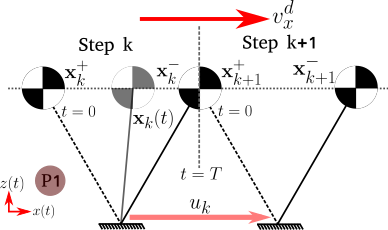}
         \caption{Forward walking}
         \label{fig:y equals x}
     \end{subfigure}
     \begin{subfigure}[b]{1\columnwidth}
         \centering
        \includegraphics[trim={0cm 0cm 0cm 0cm},clip,width=0.9\columnwidth]{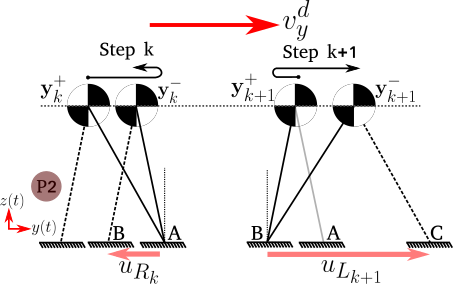}
         \caption{Lateral walking}
         \label{fig:y equals x}
     \end{subfigure}
\caption{\small{The robot motion is decomposed into two planes of motion, both decoupled motions are designed to have the swing foot to impact at time $t=T$. a) The forward walking gait with average velocity $v^d_x$ and foot placement at $u_k$ is characterized by a P1 orbit. b) The lateral walking gait with an average velocity $v^d_y$ and foot placement $u_{R_k}$ (right stance) or $u_{L_k}$ (left stance) is characterized by a P2 orbit. In both cases, the superscripts $(+)$ and $(-)$ represent the beginning of the current step, just after the impact and the end of the current step, just before the impact.}
} 
\label{fig:P1_P2_walk}
\end{figure}

Consider a sequence of walking steps realized by the biped, as shown in \figref{fig:P1_P2_walk}. For a given step $k$ we can relate the states at the beginning of the current step with the states at the end of the current step.
\begin{align}
    \mathbf{x}_k^- := \mathbf{x}(T) = M(T) \mathbf{x}_k^+, 
    \label{eq:lipsol}
\end{align}
where $\mathbf{x}_k^-$ are the pre-impact states, $\mathbf{x}_k^+$ represents the post-impact states in step $k$, and $T$ is the step duration. The equation \eqref{eq:lipsol} can be made time-independent by choosing a constant step duration. We will drop the dependency on the constant time $T$, i.e., $M(T)=M$, on future equations for simplicity.
The motion of an under-actuated robot under our LIP assumptions is completely described by setting its initial condition $\mathbf{x}_0$ and obtaining its solution through \eqref{eq:nonautonomous}. As it will be shown in the following section, the initial condition $\mathbf{x}_0$ in the LIP determines the type of motion it will describe (frontal or sagittal plane motion). 

\begin{figure}
\centering
\includegraphics[trim={0cm 0cm 0cm 0cm},clip,width=1\columnwidth]{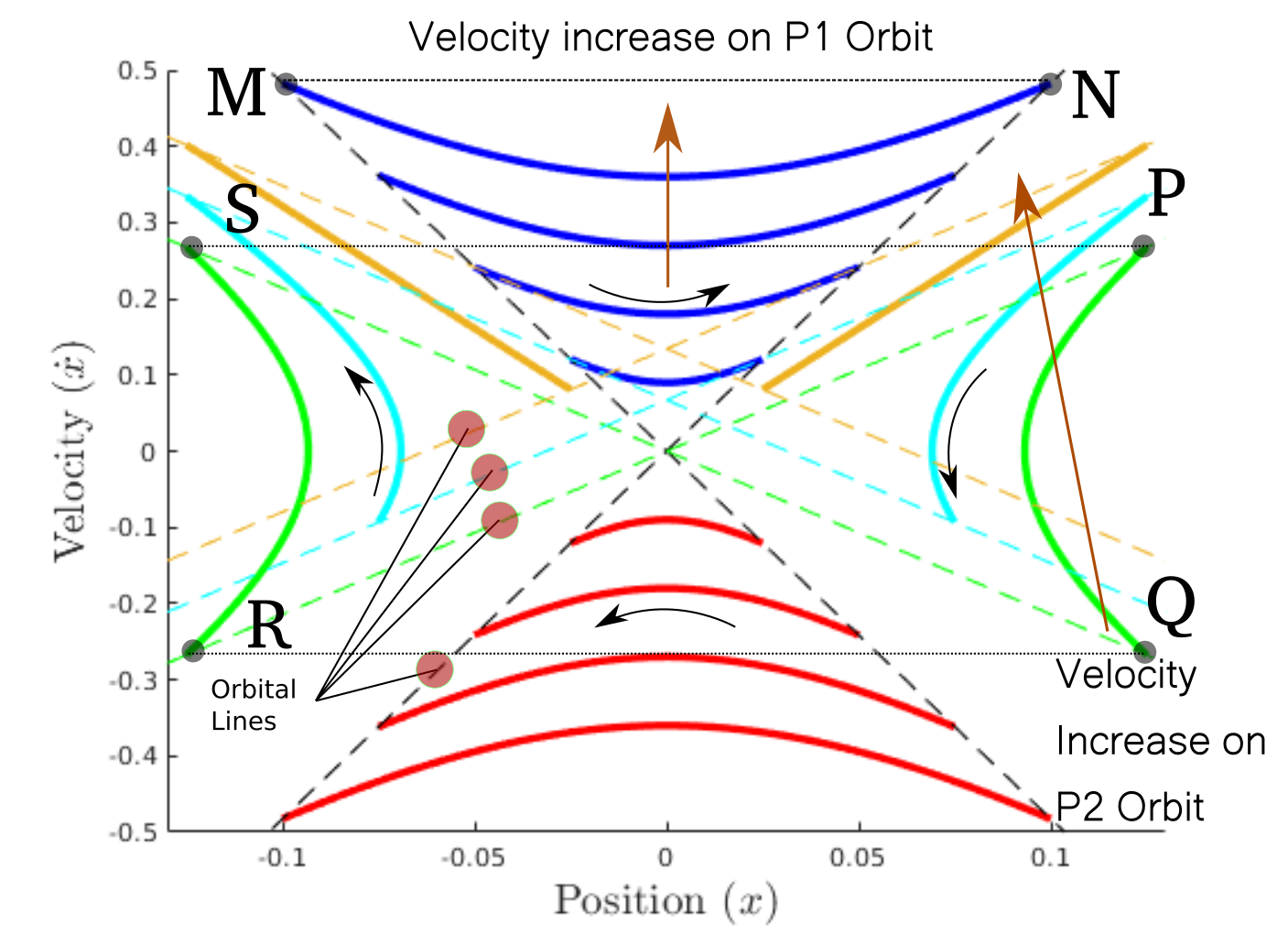}
\caption{\small{The LIP phase portrait shows different state evolution for forward velocity (P1 orbits) and lateral velocity (P2 orbits). The dashed lines represent the orbital lines, which serve as boundary of the motion given a stepping time $T$ and a desired velocity. The dotted horizontal lines represent the reset map that restarts the orbit for the next step. Note that P1 orbits are 1-step periodic, while P2 orbits are 2-step periodic.}} 
\label{fig:phase_portrait}
\end{figure}

\subsection{Phase Portrait and LIP Orbits}
\label{sec:traj}
There are two different types of phase portrait trajectories that the LIP exhibit, shown in \figref{fig:phase_portrait}. They are called P1 and P2 orbits. In the P1 orbits, we represent the forward and backward motion with the blue and red lines, respectively. On the other hand, P2 orbits are represented by the green, cyan and orange lines to represent lateral walking for different speeds.

The walking orbits are described by the phase portrait of the LIP dynamics when we show periodic solutions with a step duration $t=T$. For instance, considering the P1 orbits in \figref{fig:phase_portrait}, we can start at the point \textbf{M} and follow the LIP dynamics (blue line) for $T$ seconds until reaching the symmetric point \textbf{N}. When reaching \textbf{N} an impact occurs and the swing leg becomes the support leg. The new initial condition of the LIP depends on the location of the swing foot, if we set it such that the gait is periodic, the transition is represented by the horizontal dotted line that goes instantaneously from \textbf{N} to \textbf{M}, completing an step.
Similarly, for P2 orbits, if we start at point \textbf{P} and follow the LIP dynamics (green line) the state will reach \textbf{Q} and the swing leg will impact with the ground. After swapping legs, and assuming a periodic solution, the state becomes \textbf{R} and  moving towards \textbf{S} . Finally, after impacting with the ground again, the state returns to \textbf{P}.

The orbital lines (dashed lines in \figref{fig:phase_portrait}) represent the boundaries on which the continuous part of a periodic gait begins and ends.

\newsec{P1 Orbital Lines.}
The P1 orbital lines conform a set of initial conditions of the LIP dynamics that provides periodic solutions in $T$ seconds, i.e $p_x(T) = -p_x(0)$ and $v_x(T) = v_x(0)$. The P1 orbital lines are defined by the equation:
\begin{align}
    v_x &= \pm \lambda \coth(\frac{T \lambda}{2}) p_x =\pm \sigma_1 p_x, 
\end{align}
where $\sigma_1$ is the slope of the P1 orbital lines and determined solely by the step duration, $T$.

\newsec{P2 Orbital Lines.}
Likewise, the green, cyan and orange lines in \figref{fig:phase_portrait} represent P2 orbits for different positive lateral velocities ($v^d_y$). The P2 orbital lines are characterized by the desired lateral velocity, given as
\begin{align}
    v_y &= \pm \sigma_2 p_y + d_2,
\end{align}
where $\sigma_2 = \lambda \tanh(\frac{T \lambda}{2})$ is the slope and 
\begin{align}
    d_2 &= \frac{\lambda^2 \sech^2(\frac{\lambda T}{2}) T v^d_y}{2\sigma_2},    
\end{align}
is the offset. 
A periodic gait can be synthesized from the phase portrait by finding a periodic motion with the desired forward and lateral velocities as shown in \figref{fig:P1_P2_walk}.
A detailed computation of the orbital lines, including the case where double support phase exists can be found in \cite{xiong2020dynamic}.

\section{Foot placement control}
\label{sec:control}
A foot placement controller can be designed with the objective to drive the actual robot states to the ideal states of the LIP just before impact. We use an auxiliary control input---the swing foot location---in our design.

\subsection{Target LIP states}
For P1 orbits, the ideal LIP states just before impact ($\mathbf{x}^*$) are located on the line $v_x = +\sigma_1 p_x$, 
\begin{align}
    \mathbf{x}^* = \begin{bmatrix} 1 & \sigma_1 \end{bmatrix}^T \frac{v^d_x T}{2}. \label{eq:P1x}
\end{align}
To keep a periodic solution the ideal swing foot location is designed as $u_x = u^*$ where,
\begin{align}
    u^* = v^d_x T.
\end{align}

In the P2 orbits, $\mathbf{y}^*$ can represent the swing foot with respect to to left foot ($\mathbf{y}^*_L$) or right foot ($\mathbf{y}^*_R$). The ideal foot placement, $u_y = u^*_{L/R}$ is required to compute the ideal states,
\begin{align}
    \mathbf{y}^*_{L/R} &= \begin{bmatrix} u^*_{L/R}/2 \\
    \sigma_2 u^*_{L/R} + d_2 \end{bmatrix} \label{eq:P2x}
\end{align}
Note that $u^*_{L/R}$ has many solutions because we have only one constraint: $u^*_L + u^*_R = v^d_y T$. We need to define one of them, ($u^*_L$ for instance) to calculate the other one.

Next, we obtain the discrete relationship that map two consecutive walking steps using the commanded foot placement. For simplicity, but without loss of generality, we drop the subscripts ($x$ or $y$) of the LIP and represent the states as $p$ and $v$, with $\mathbf{x} = (p,v)$ and control $u$.

\subsection{Step to step dynamics}
Consider a domain transition from the end of one cycle (just before impact) to the beginning of the next one (just after impact), as shown in \figref{fig:P1_P2_walk} part a),
\begin{align}
    p^{+}_{k+1} &= -u_{k}, \\
    v_{k+1}^{+} &= v_k^{-},
\end{align}
which can be represented in an affine form by:
\begin{align}
    \mathbf{x}_{k+1}^{+} &= \begin{bmatrix} 1 & 0 \\ 0 & 1 \end{bmatrix} \begin{bmatrix} x^{-} \\ \dot{x}^{-} \end{bmatrix}_k + \begin{bmatrix} -1 \\ 0 \end{bmatrix} u_k = a \mathbf{x}_k^{-} + b u_k.
    \label{eq:lipdiscrete}
\end{align}
Combining \eqref{eq:lipsol} and \eqref{eq:lipdiscrete}, we can obtain the step-to-step dynamics of the LIP model:
\begin{align}
     \mathbf{x}_{k+1}^{-} = M (a \mathbf{x}_k^{-} + b u_k) = A\mathbf{x}_k^{-} + B u_k.
    \label{eq:ss_ideal}
\end{align}
We drop the (-) in the following discussion for simplicity.  The step-to-step dynamics in \eqref{eq:ss_ideal} determines the linear mapping of before-impact states between two consecutive steps.
This system accepts a linear controller of the form: 
\begin{align}
    u_k = u^*_k + K(\mathbf{x}_k - \mathbf{x}^*_{k}),  
\end{align}
where $u^*_k$ is the ideal step length provided by the target LIP state and $K$ is a linear gain.

\subsection{Error Dynamics}
We define the error at step $k$, as the difference between the actual LIP states of the robot ($\mathbf{x}_{k}$) and its ideal value ($\mathbf{x}^*_k$), given in \eqref{eq:P1x} for P1 orbits and in \eqref{eq:P2x} for P2 orbits, $e_k = \mathbf{x}_{k} - \mathbf{x}^*_k$.
Also consider the error in model approximation $\xi$ integrated in \eqref{eq:ss_ideal}:
\begin{align}
    \mathbf{x}_{k+1} = A \mathbf{x}_k + B u_k + \xi. \label{eq:ss_actual}
\end{align}
We can compute the error propagation in the next step $k+1$ and relate it with the states and input of the current step $k$.
\begin{align}
    e_{k+1} &= \mathbf{x}_{k+1} - \mathbf{x}^*_{k+1} \nonumber \\
     &= (A+BK)e_k + \xi. \label{eq:error1}
\end{align}
Under a perfect match between the LIP and the real robot ($\xi=0$) we choose $K$ such that the error $e_{k+2}$ becomes zero,
\begin{align}
    (A+BK)^2 &= 0.
\end{align}
Considering the system affected by modelling errors in \eqref{eq:ss_actual}, we observe that we carry the errors for $N > 2$ steps.
\begin{align}
    e_{k+N} &= (A+BK+I) \xi. \label{eq:errorN}
\end{align}
This shows that the modelling error $\xi$ generates a bounded error in the robot states away from $\mathbf{x}^*$.

\subsection{Neural Adaptive Regulator}
As observed, the template model generates a persistent error on the states due to model mismatch. To improve the controller action, we propose to add a feed-forward term that captures the un-modeled dynamics based on a non-supervised neural network, as detailed in \cite{thakkar2021adaptive}:
\begin{align}
    u_k = u^*_k + K(\mathbf{x}_k - \mathbf{x}^*_{k}) + \phi(\mathbf{x}_k, v^d_x, v^d_y),
\end{align}
where $\phi$ is the function representing a two layer neural network
\begin{align}
    \phi(\mathbf{x}_k, v^d_x, v^d_y) = \sigma \left( W^T \sigma \left(V^T  \begin{bmatrix}  \mathbf{x}_k \\ v^d_x \\ v^d_y \end{bmatrix} \right) \right).
\end{align}


The input layer weights $\hat{V}$ are initialized with a normal distribution and $\hat{W}$ are the output layer weights that are learned in real-time, using gradient descent delta-rule. The weight update is computed as, 
\begin{align}
    E &= (\mathbf{x}_k-\mathbf{x}^*_k)^2, \\
    \label{weight_update}
    \Delta w_{i,j} &= -\gamma E_j \sigma_i,
\end{align}
where, $w_{i,j}$ is the weight from the $i^{th}$ hidden neuron to the $j^{th}$
output, $E_j$ is the error signal for the $j^{th}$ output, $\sigma_i$ is the activation function evaluated at the $i$-th output and $\gamma$ is the learning rate chosen as
$\gamma = 1e{-4}$.

The objective of the neural network is to minimize the error on the state tracking of the LIP model by minimizing the error between $\mathbf{x}_k$ and $\dot{\mathbf{x}}^*_k$ as a feed-forward term that compensates for the modeling error $\xi$.
\section{QP-based Resolved Inverse Kinematics}
\label{sec:ik}
The foot placement controller returns the desired location of the swing foot for the current step, $u_x$ and $u_y$, in the $x$ and $y$ direction respectively. Then, we construct a set of polynomials that provides a trajectory that connects the current swing foot location to its desired location at time $T,$ furthermore, such trajectory generates human-like walking patterns with adequate foot clearance and smoothness. 

\subsubsection{Task space trajectories}
We employ B\'ezier Polynomials that connect the initial configuration of the CoM and swing foot frame with one where $u_x$ and $u_y$ are realized, while keeping the torso frame vertical and the swing foot horizontal with the floor (in both cases the rotation matrix with respect to support foot is the identity).

Since the LIP is considered as point-foot, we allow the ankle (sagittal and frontal direction) of the actual robot to be not-actuated, making the CoM in $x$ and $y$ directions to be free, but we impose a constant height in the $z$ direction. 
Consequently, the reference for the CoM frame, oriented as the pelvis frame, are:
\begin{align}
        p_{CoM}^z(t) &= z_0, \\
        R_{CoM}(t) &= I_3,
\end{align}
where $p_{CoM}^z(t)$ is the center of mass height of the robot with respect to the support foot, $z_0$ is a constant height reference (see \figref{fig:lip_digit}) and $R_{CoM}(t)$ is the rotation matrix of the center of mass frame with respect to the support foot, and $I_3$ is an identity matrix.

In the case of the swing foot, we construct references for position and orientation in {$x,y,z$} coordinates,
\begin{align}
    p_{sw}(t) &= \sum_{i=0}^{n} { n \choose i } P_i (1-t)^{n-i}t^i, \\
    R_{sw}(t) &= I_3,
\end{align}
where $p_{sw}(t)$ represents the spatial position of the swing foot, $n$ is order of the B\'ezier polynomials and $P_i$ are the control points. $R_{sw}(t)$ represent the rotation matrix of the swing foot. The polynomials are created to be coincident with the initial swing position just after the impact and the desired swing foot location just before the impact. 
\begin{align}
    p_{sw}(0) &= p_0, \\
    p_{sw}(T) &= \begin{bmatrix} u_x & u_y & -z_0\end{bmatrix}^T
\end{align}
where $p_0$ is the initial swing foot position at the beginning of a walking step.


\subsection{QP based Inverse Kinematics}
A standard way to follow the trajectories is through a model-based controller running in a QP program. Even when the modeling is imperfect, a system identification algorithm, or a model based adaptive controller \cite{nguyen20151, sun2021online} can be applied. A simpler method can be designed to use the kinematics information of the system to provide joint-space reference trajectories via a QP-based Inverse Kinematics problem that runs in real time \cite{suleiman2015infeasibility, kanoun2012real, nakanishi2008operational}.


In this paper, we solve the optimization problem of the full robotic model with joint coordinates $(q, \dot{q}) \in TQ$. The objective function minimizes the error in position and orientation of the CoM and swing foot frames. 
 Our decision variable is the desired joint velocity $\dot{q}^d$, which we use to constrain the maximum joint velocity, the joint position limits and to specify some joints as passive. The optimization is shown below,
\begin{align}
    \displaystyle \min_{\dot{q}^d} \hspace{1em} & || J_{CoM} \dot{q}^d - T_{CoM}(t) || + || J_{sw} \dot{q}^d - T_{sw}(t) || \\
    \st & \dot{q}^d \in [\dot{q}_{min}, \dot{q}_{max}] \nonumber \\
    & q^d =q + \delta_t \dot{q}^d \in [q_{min}, q_{max}] \nonumber \\
    & \dot{q}^d_{\text{passive joints}} = \dot{q}_{\text{passive joints}}, \nonumber
\end{align}
    with each objective $T_{CoM}(t), T_{sw}(t)$ with $m\in\{CoM, sw\}$ defined as a reference velocity plus a correction term:
\begin{align}
    T_m(t) = \begin{bmatrix} v_m^d(t) + K_p (p_m^d(t) - p_m) \\ \omega_m^d(t) - K_{\omega} e_m(t) \end{bmatrix}, \nonumber
\end{align}
    where $e_m(t) = \eta_m^d(t) \epsilon_m - \eta_m \epsilon_m^d(t) + \epsilon_m^d(t) \times \epsilon_m$ represents the orientation error. Consider the desired the quaternion as $\alpha_m^d(t) = \begin{bmatrix} \eta_m^d(t), \mathbf{\epsilon}_m^d(t) \end{bmatrix}$ and the current quaternion as $\alpha_m = \begin{bmatrix} \eta_m, \mathbf{\epsilon_m} \end{bmatrix}$. Each quaternion is composed by a scalar part $\eta$ and a vector part $\epsilon$. $J_{CoM}$ and $J_{sw}$ are the respective jacobian matrices for the CoM and swing foot frame. The two gain matrices $K_p$ and $K_{\omega}$ can be tuned to reduce the errors on the desired tasks. 
    A detailed analysis of the objective vectors $T_{CoM}$ and $T_{sw}$ and the QP can be found in \cite{nakanishi2008operational}.
    
\begin{figure}
\centering
\vspace{2mm}
\includegraphics[trim={0cm 0cm 0cm 0cm},clip,width=1\columnwidth]{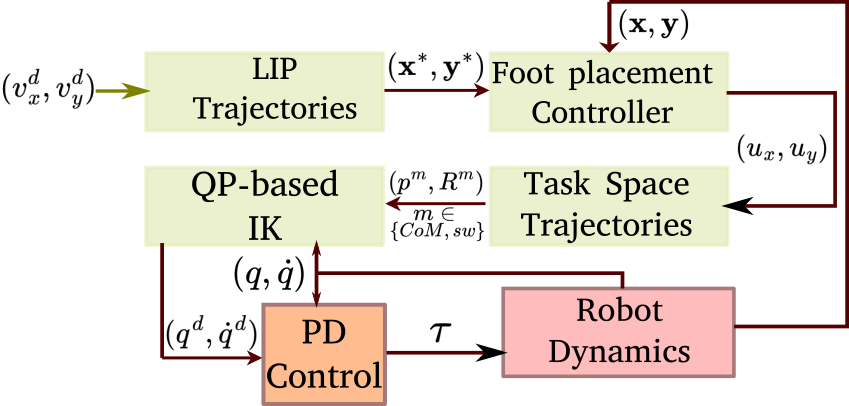}
\caption{\small{The resolved motion framework starts with the ideal LIP trajectories that provides target states for the robot $(\mathbf{x}^*, \mathbf{y}^*)$ at the end of the current gait, they are compared with the estimation of the states at the end of the current gait $(\mathbf{x}, \mathbf{y})$, then the foot placement controller provides desired foot locations $(u_x,u_y)$. These locations are transformed into Task space references $(p_m, R_m), m \in \{CoM,sw\}$ and transformed into joint space trajectories $(q^d, \dot{q}^d)$ through the QP-based inverse kinematics which are tracked with a joint-level PD controller. }} 
\label{fig:block_diagram}
\end{figure}

We resolve the reference joint trajectories in real time by integrating the solution of the QP program, $q^d = q + \dot{q}^d \delta_t$, where $\delta_t$ is the time interval of the control loop set at $1$ ms. The interaction between foot placement control, trajectory generation and inverse kinematics can be seen in \figref{fig:block_diagram}.

\section{Results}
\label{sec:results}

The proposed method is validated in both simulation and hardware experiments, using the same C++ code and parameters, including control gains, neural network settings and stepping gains. A video is included with the procedure and the simulation and experimental results \footnote{\url{https://youtu.be/7ym2gm3XiOE}}.

\subsection{Simulation Results}
The simulation software is provided by Agility Robotics, which is specially tailored to provide a realistic representation of Digit. The provided low-level API handles the interaction with the robot in simulation and hardware implementation. We use the simulation to ensure proper robot behavior and test the control framework.  

\begin{figure}
\centering
\vspace{2mm}
\includegraphics[trim={0cm 0cm 0cm 0cm},clip,width=1\columnwidth]{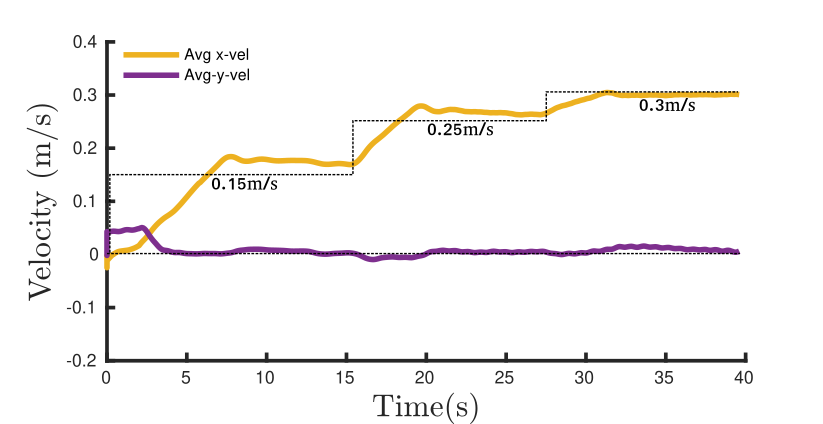}
\caption{\small{Digit achieving several commanded velocities $v_x = \{0, 0.15, 0.25, 0.3\}$ and $v_y=0$ }} 
\label{fig:trackingvel_sim}
\end{figure}

The first test consisted on commanding different target forward velocities (\figref{fig:trackingvel_sim}). With each new velocity, the robot recomputes the desired orbit on the fly, generating a new placement for the swing foot. We can observe that Digit achieves velocities up to 0.3 m/s in simulation and on average.


\begin{figure}
\centering
\includegraphics[trim={1cm 0.8cm 1.5cm 0.5cm},clip,width=1\columnwidth]{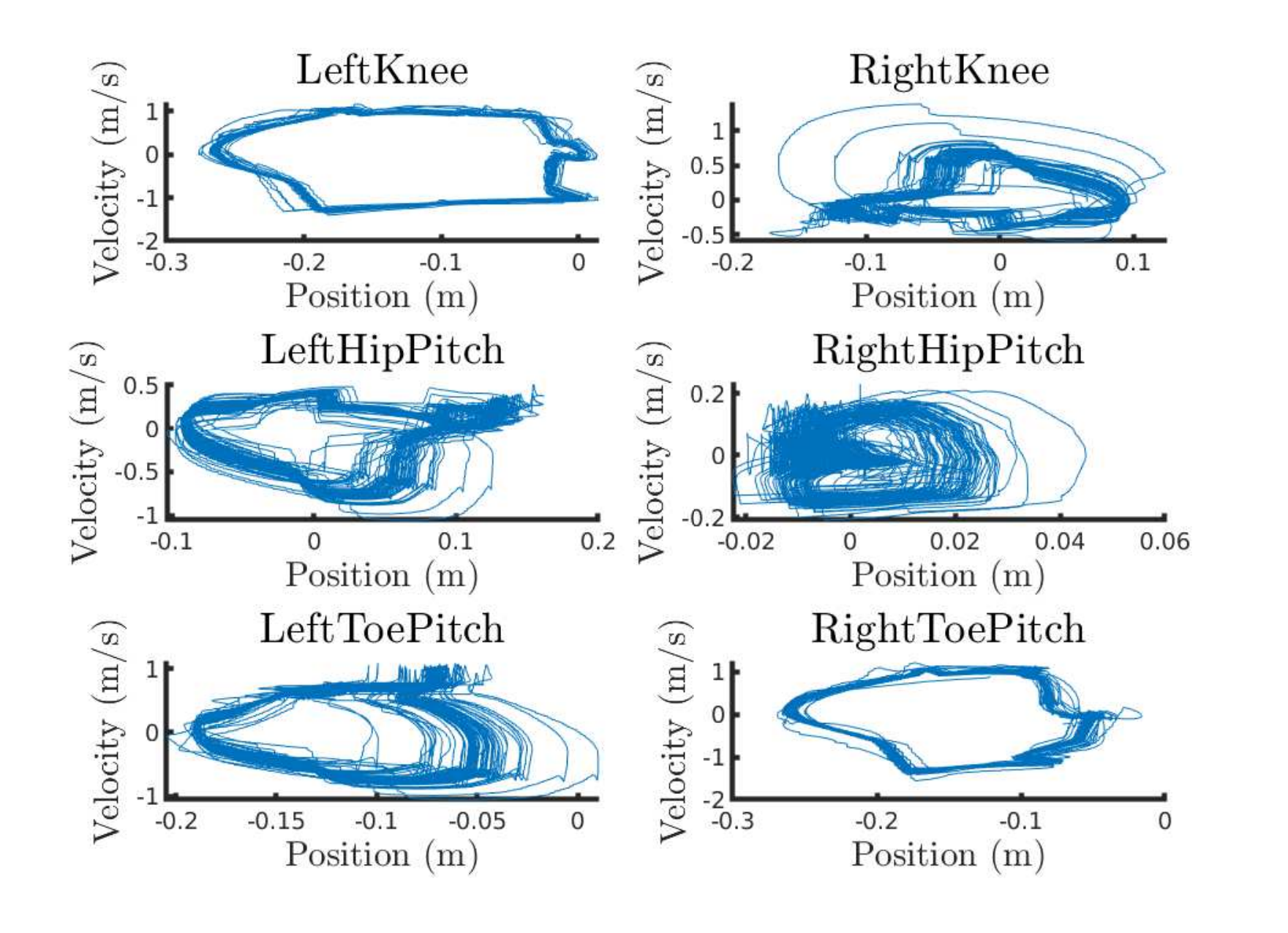}
\caption{\small{Phase portrait of selected joints during walking gait at 0.1 m/s in simulation.}} 
\label{fig:phasep_sim}
\end{figure}

To check for periodicity, we set a constant velocity to the robot and we plot the phase-portrait with the joints. We observe a periodic orbit as shown in \figref{fig:phasep_sim}.

\begin{figure}
\centering
\includegraphics[trim={1.2cm 0cm 1.8cm 0cm},clip,width=1\columnwidth]{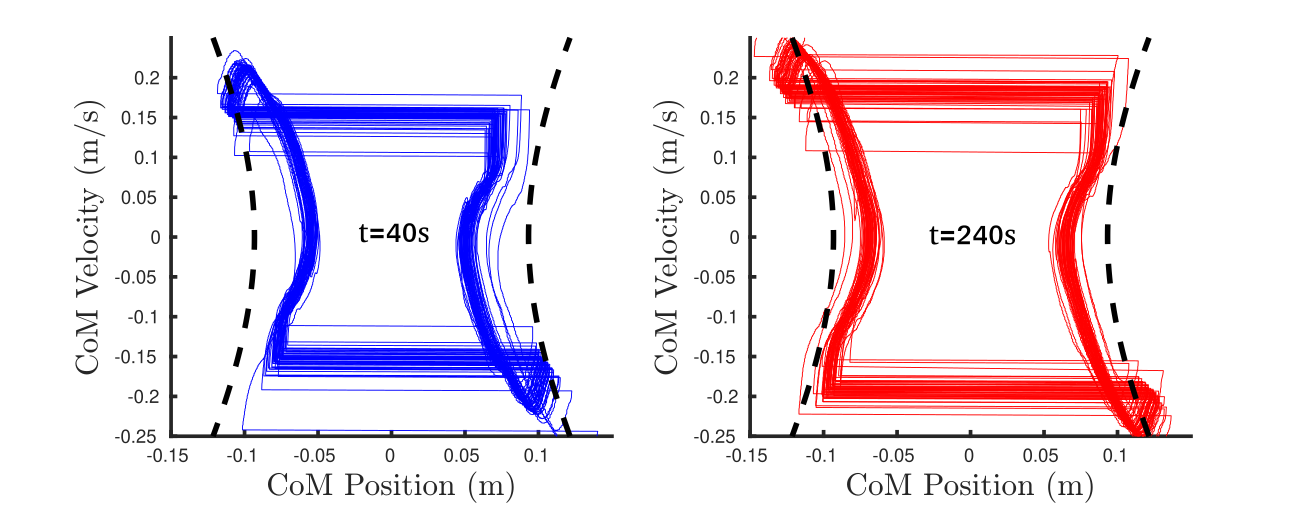}
\caption{\small{Lateral states for $v_x = 0.1$ m/s, during simulation. At 40 s, the LIP based controller cannot has an average error of ||e||=0.1 and after learning a regulation at 240s it decreases to ||e||=0.06.}} 
\label{fig:lip_phasep_sim}
\end{figure}

The effect of the neural adaptive controller is seen in \figref{fig:lip_phasep_sim}, we observed that the neural feed-forward term drives the orbit from an state error $||e||=0.1$ at 40s to $||e||=0.06$ at about 240s.

\subsection{Hardware Results}
We conducted walking experiments on a smooth flat surface in a laboratory setting and on a treadmill without inclination. The controller is executed on a host computer outside the robot's mainboard, and the commands are transmitted through an Ethernet cable at 1KHz.

\begin{figure}
\centering
\vspace{7mm}
\includegraphics[trim={0cm 0cm 0cm 0cm},clip,width=1\columnwidth]{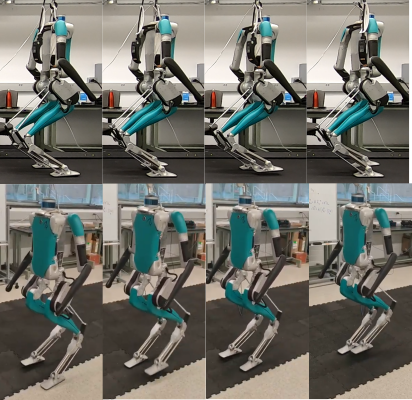}
\caption{\small{Snapshots of the gait for forward walking (above) and lateral walking (below) on flat ground.}} 
\label{fig:lateral_hw}
\end{figure}

\begin{figure}
\centering
\includegraphics[trim={0cm 0cm 0cm 0cm},clip,width=1\columnwidth]{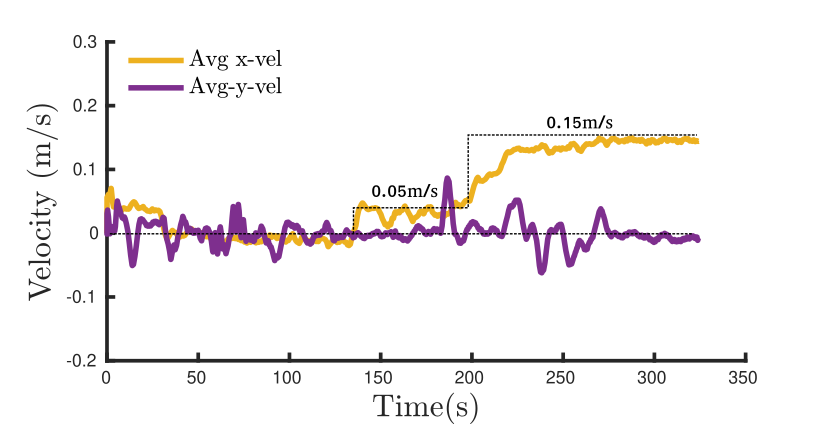}
\caption{\small{Digit achieved a velocity of $v^d_x = 0.15$ m/s during the experiments on the treadmill with $v^d_y = 0$ m/s. }} 
\label{fig:trackingvel_hw}
\end{figure}

\begin{figure}
\centering
\includegraphics[trim={1cm 0.8cm 1.5cm 0.5cm},clip,width=1\columnwidth]{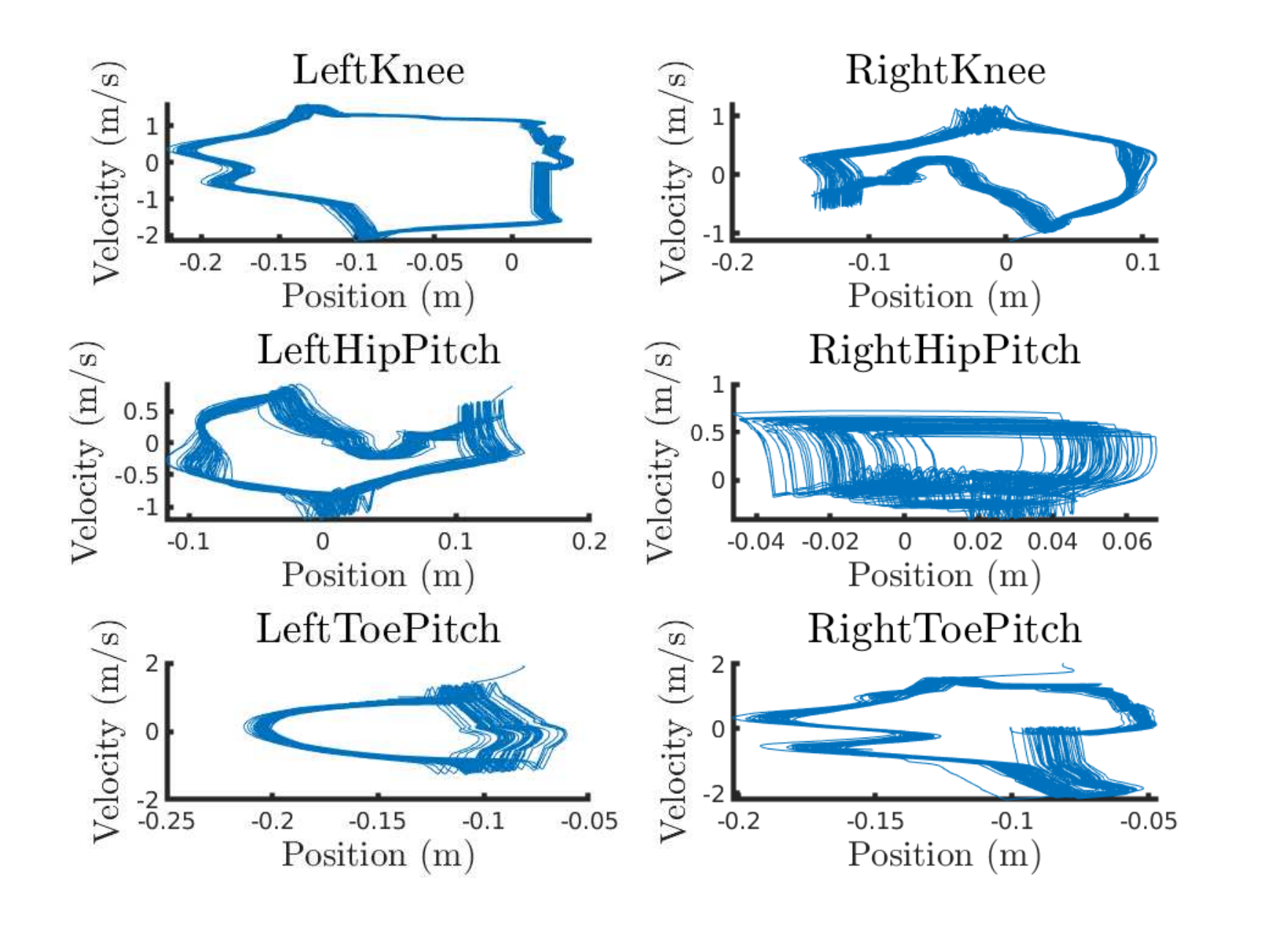}
\caption{\small{Phase portrait of selected joints during walking gait at 0.1 m/s in the robot hardware experiment.}} 
\label{fig:phasep_hw}
\end{figure}

We tested lateral and forward walking, and their respective snapshots can be observed in \figref{fig:lateral_hw}. We let the robot walk for about 6 min on the treadmill with velocity increases up to 0.15 m/s. The velocity profile of the gait is observed in \figref{fig:trackingvel_hw}, while the phase portrait of the sagittal plane joints after reaching 0.15 m/s is shown in \figref{fig:phasep_hw} indicating the periodicity of the resulting gait.

\begin{figure}
\centering
\includegraphics[trim={1.2cm 0cm 1.8cm 0cm},clip,width=1\columnwidth]{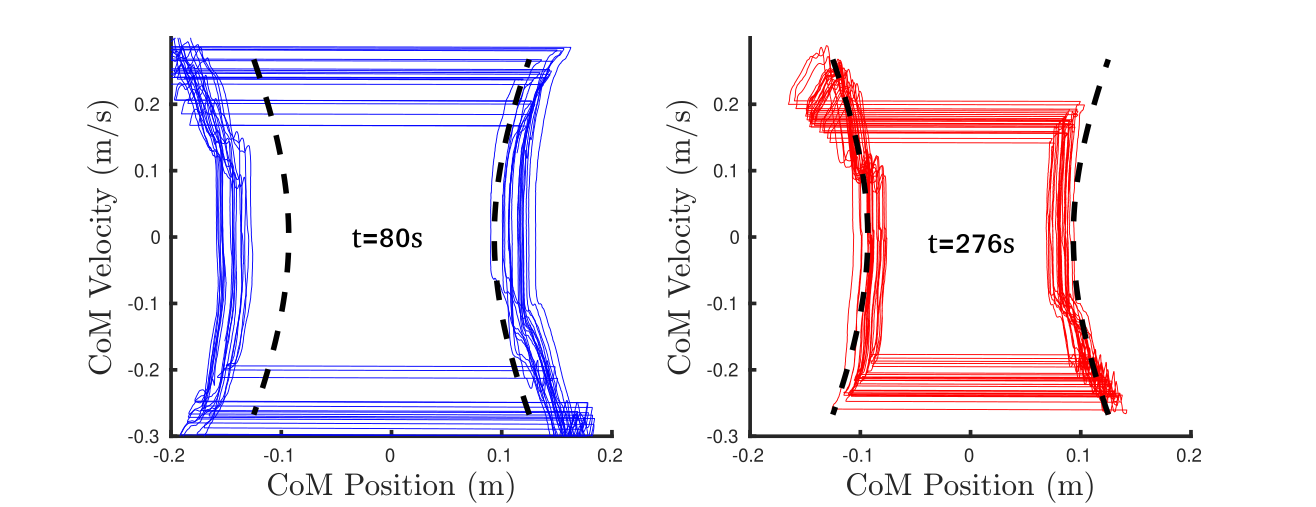}
\caption{\small{Lateral states for $v_x = 0.1$ m/s, during the hardware experiments. At 85s, the LIP based controller cannot drive the robot states to the LIP reference values until the neural regulator learns a feed-forward term at 276s.}} 
\label{fig:lip_phasep_hw}
\end{figure}

Similarly to the simulation case, we set a constant forward velocity $v_x = 0.1m/s$  \figref{fig:lip_phasep_hw}, shows that the initial states have an error of $||e||=0.12$ at 80s, while the learned regulator leads to $||e||=0.08$ at 276s.

\section{Conclusions}
We present a LIP based control method that generates swing foot locations that produce stable walking for 3D robots with passive ankles. We realize that model mismatch can generate a persistent error on a LIP based linear controller and that by using a neural adaptive regulator we can decrease this error. Additionally, we use an online QP-based inverse kinematics problem that can be solved in real time and provides joint level references to be tracked by a PD controller. This allows a fast method that is independent of the dynamics of the robot. Finally, the framework realize stable and periodic walking for Digit in both, simulation and hardware experiments, and can track different forward and lateral velocities on the fly.

\bibliographystyle{IEEEtran}
\bibliography{references.bib}

\begin{thebibliography}{10}
\providecommand{\url}[1]{#1}
\csname url@rmstyle\endcsname
\providecommand{\newblock}{\relax}
\providecommand{\bibinfo}[2]{#2}
\providecommand\BIBentrySTDinterwordspacing{\spaceskip=0pt\relax}
\providecommand\BIBentryALTinterwordstretchfactor{4}
\providecommand\BIBentryALTinterwordspacing{\spaceskip=\fontdimen2\font plus
\BIBentryALTinterwordstretchfactor\fontdimen3\font minus
  \fontdimen4\font\relax}
\providecommand\BIBforeignlanguage[2]{{%
\expandafter\ifx\csname l@#1\endcsname\relax
\typeout{** WARNING: IEEEtran.bst: No hyphenation pattern has been}%
\typeout{** loaded for the language `#1'. Using the pattern for}%
\typeout{** the default language instead.}%
\else
\language=\csname l@#1\endcsname
\fi
#2}}

\bibitem{englsberger2015three}
J.~Englsberger, C.~Ott, and A.~Albu-Sch{\"a}ffer, ``Three-dimensional bipedal
  walking control based on divergent component of motion,'' \emph{Ieee
  transactions on robotics}, vol.~31, no.~2, pp. 355--368, 2015.

\bibitem{griffin2016model}
R.~J. Griffin and A.~Leonessa, ``Model predictive control for dynamic footstep
  adjustment using the divergent component of motion,'' in \emph{2016 {IEEE}
  {International} {Conference} on {Robotics} and {Automation} ({ICRA})}, May
  2016, pp. 1763--1768.

\bibitem{kajita20013d}
S.~Kajita, F.~Kanehiro, K.~Kaneko, K.~Yokoi, and H.~Hirukawa, ``The {3D} linear
  inverted pendulum model: a simple modeling for a biped walking pattern
  generation,'' in \emph{Proceedings of the IEEE/RSJ International Conference
  on Intelligent Robots and Systems}, 2001, pp. 239--246.

\bibitem{pratt2012capturability}
J.~Pratt, T.~Koolen, T.~de~Boer, J.~Rebula, S.~Cotton, J.~Carff, M.~Johnson,
  and P.~Neuhaus, ``Capturability-based analysis and control of legged
  locomotion, part 2: application to m2v2, a lower-body humanoid,'' \emph{The
  International Journal of Robotics Research}, vol.~31, no.~10, pp. 1117--1133,
  Aug. 2012.

\bibitem{goswami1996limit}
A.~Goswami, B.~Espiau, and A.~Keramane, ``Limit cycles and their stability in a
  passive bipedal gait,'' in \emph{Proceedings of IEEE international conference
  on robotics and automation}, vol.~1.\hskip 1em plus 0.5em minus 0.4em\relax
  IEEE, 1996, pp. 246--251.

\bibitem{hereid20163d}
A.~Hereid, E.~A. Cousineau, C.~M. Hubicki, and A.~D. Ames, ``{3D} dynamic
  walking with underactuated humanoid robots: a direct collocation framework
  for optimizing hybrid zero dynamics,'' in \emph{Proc. IEEE Int. Conf.
  Robotics and Automation (ICRA)}.\hskip 1em plus 0.5em minus 0.4em\relax
  Stockholm, Sweden: IEEE, May 2016, pp. 1447--1454.

\bibitem{da2019combining}
X.~Da and J.~Grizzle, ``Combining trajectory optimization, supervised machine
  learning, and model structure for mitigating the curse of dimensionality in
  the control of bipedal robots,'' \emph{The International Journal of Robotics
  Research}, vol.~38, no.~9, pp. 1063--1097, jul 2019.

\bibitem{gong2021one}
Y.~Gong and J.~Grizzle, ``One-step ahead prediction of angular momentum about
  the contact point for control of bipedal locomotion: Validation in a
  lip-inspired controller,'' in \emph{2021 IEEE International Conference on
  Robotics and Automation (ICRA)}, 2021, pp. 2832--2838.

\bibitem{xiong20213d}
X.~Xiong and A.~Ames, ``3d underactuated bipedal walking via h-lip based gait
  synthesis and stepping stabilization,'' \emph{arXiv preprint
  arXiv:2101.09588}, 2021.

\bibitem{xiong2020dynamic}
X.~Xiong and A.~D. Ames, ``Dynamic and versatile humanoid walking via embedding
  3d actuated slip model with hybrid lip based stepping,'' \emph{IEEE Robotics
  and Automation Letters}, vol.~5, no.~4, pp. 6286--6293, 2020.

\bibitem{xie2021glide}
Z.~Xie, X.~Da, B.~Babich, A.~Garg, and M.~van~de Panne, ``Glide: Generalizable
  quadrupedal locomotion in diverse environments with a centroidal model,''
  \emph{arXiv preprint arXiv:2104.09771}, 2021.

\bibitem{castillo2020hybrid}
G.~Castillo, B.~Weng, W.~Zhang, and A.~Hereid, ``Hybrid zero dynamics inspired
  feedback control policy design for 3d bipedal locomotion using reinforcement
  learning,'' in \emph{IEEE International Conference on Robotics and Automation
  (ICRA)}.\hskip 1em plus 0.5em minus 0.4em\relax Paris, France: IEEE, May
  2020.

\bibitem{paredes2020dynamic}
V.~Paredes and A.~Hereid, ``Dynamic locomotion of a lower-limb exoskeleton
  through virtual constraints based zmp regulation,'' in \emph{Dynamic Systems
  and Control Conference}, vol. 84270.\hskip 1em plus 0.5em minus 0.4em\relax
  American Society of Mechanical Engineers, 2020, p. V001T14A001.

\bibitem{raibert1986legged}
M.~H. Raibert \emph{et~al.}, \emph{Legged robots that balance}.\hskip 1em plus
  0.5em minus 0.4em\relax MIT press Cambridge, MA, 1986, vol.~3.

\bibitem{teng2021safety}
S.~Teng, Y.~Gong, J.~W. Grizzle, and M.~Ghaffari, ``Toward safety-aware
  informative motion planning for legged robots,'' \emph{arXiv:2103.14252
  [cs]}, Mar. 2021.

\bibitem{nguyen20151}
Q.~Nguyen and K.~Sreenath, ``L 1 adaptive control for bipedal robots with
  control lyapunov function based quadratic programs,'' in \emph{2015 American
  Control Conference (ACC)}.\hskip 1em plus 0.5em minus 0.4em\relax IEEE, 2015,
  pp. 862--867.

\bibitem{sun2021online}
Y.~Sun, W.~L. Ubellacker, W.-L. Ma, X.~Zhang, C.~Wang, N.~V. Csomay-Shanklin,
  M.~Tomizuka, K.~Sreenath, and A.~D. Ames, ``Online learning of unknown
  dynamics for model-based controllers in legged locomotion,'' \emph{IEEE
  Robotics and Automation Letters}, vol.~6, no.~4, pp. 8442--8449, 2021.

\bibitem{gong2021zero}
Y.~Gong and J.~Grizzle, ``Zero {Dynamics}, {Pendulum} {Models}, and {Angular}
  {Momentum} in {Feedback} {Control} of {Bipedal} {Locomotion},''
  \emph{arXiv:2105.08170 [cs, eess]}, May 2021.

\bibitem{thakkar2021adaptive}
K.~Thakkar, V.~Paredes, and A.~Hereid, ``Adaptive feedback regulator for
  powered lower-limb exoskeleton under model uncertainty,'' \emph{arXiv
  preprint arXiv:2104.11775}, 2021.

\bibitem{suleiman2015infeasibility}
W.~Suleiman, F.~Kanehiro, and E.~Yoshida, ``Infeasibility-free inverse
  kinematics method,'' in \emph{2015 {IEEE}/{SICE} International Symposium on
  System Integration ({SII})}, IEEE.\hskip 1em plus 0.5em minus 0.4em\relax
  {IEEE}, dec 2015, pp. 307--312.

\bibitem{kanoun2012real}
O.~Kanoun, ``Real-time prioritized kinematic control under inequality
  constraints for redundant manipulators,'' in \emph{Robotics: Science and
  Systems}, vol.~7, 2012, p. 145.

\bibitem{nakanishi2008operational}
J.~Nakanishi, R.~Cory, M.~Mistry, J.~Peters, and S.~Schaal, ``Operational space
  control: A theoretical and empirical comparison,'' \emph{The International
  Journal of Robotics Research}, vol.~27, no.~6, pp. 737--757, 2008.

\end{thebibliography}

\end{document}